%% file: main.tex
\documentclass[11pt,a4paper]{article}
\usepackage{times,latexsym}
\usepackage{url}
\usepackage[T1]{fontenc}

\usepackage{microtype}
\usepackage{url}
\usepackage[utf8]{inputenc}
\usepackage[T1]{fontenc}
\usepackage{color,soul}
\setul{0.45ex}{0.25ex}
\usepackage{graphicx}
\usepackage{booktabs}
\usepackage{multirow, makecell}
\usepackage{tikz}
\usepackage{graphicx}
\usepackage{array}
\usepackage{pgfplots}
\usepackage{algorithm}
\usepackage{textcomp}
\usepackage{xcolor}
\usepackage{times}
\usepackage{latexsym}
\usepackage{amsmath}
\usepackage{amsfonts}
\usepackage{amssymb}
\usepackage[noend]{algpseudocode}
\usepackage{lipsum}
\usepackage{multirow}
\usepackage{framed}
\usepackage{graphicx}
\usepackage{pifont}
\usepackage{longtable}
\usepackage{tikz}
\usepackage{booktabs}
\usepackage[all]{nowidow}
\usepackage{amssymb}
\usepackage{enumitem}
\usepackage{color,soul}
\usepackage{bm}
\usepackage{arydshln}
\usepackage{xspace}
\usepackage{siunitx}
\usepackage{tabularx}
\usepackage{caption}
\usepackage{subcaption}
\usepackage{pifont}

\pgfplotsset{compat=1.17}
\usepackage[acceptedWithA]{tacl2018v2}

\usepackage{xspace,mfirstuc,tabulary}

\newif\iftaclinstructions
\taclinstructionsfalse 
\iftaclinstructions
\renewcommand{\confidential}{}
\renewcommand{\anonsubtext}{(No author info supplied here, for consistency with
TACL-submission anonymization requirements)}
\newcommand{\instr}
\fi

\iftaclpubformat 

\else

\fi

\newlength{\mysize}

\newif\ifcomments
\commentstrue
\ifcomments
    \providecommand{\eric}[2][]{{\protect\color{red}{[Eric:\textbf{#1} #2]}}}
    \providecommand{\sebastian}[2][]{{\protect\color{orange}{[SR:\textbf{#1} #2]}}}
    \providecommand{\sameer}[2][]{{\protect\color{blue}{[Sameer:\textbf{#1} #2]}}}
    \providecommand{\rob}[2][]{{\protect\color{violet}{[RL:\textbf{#1} #2]}}}
    \providecommand{\ivana}[2][]{{\protect\color{teal}{[Ivana:\textbf{#1} #2]}}}
    \providecommand{\fabio}[2][]{{\protect\color{violet}{[FP:\textbf{#1} #2]}}}
\else
    \providecommand{\eric}[2][]{}
    \providecommand{\sebastian}[2][]{}
    \providecommand{\sameer}[2][]{}
    \providecommand{\rob}[2][]{}
    \providecommand{\ivana}[2][]{}
    \providecommand{\fabio}[2][]{}
\fi

\newcommand{\mask}{\texttt{[MASK]}\xspace}
\newcommand{\cls}{\texttt{[CLS]}\xspace}
\newcommand{\autoprompt}{\textsc{AutoPrompt}\xspace}

\newcommand{\wins}{\# Wins\xspace}
\newcommand{\STAB}[1]{\begin{tabular}{@{}c@{}}#1\end{tabular}}  

\newcommand{\cmark}{\ding{51}}%
\newcommand{\xmark}{\ding{55}}%

\definecolor{dark}{HTML}{264653}
\definecolor{med}{HTML}{2A9D8F}
\definecolor{light}{HTML}{E9C46A}

\title{
    Cutting Down on Prompts and Parameters: \\ Simple Few-Shot Learning with Language Models
}

\author{
    \bf Robert L. Logan IV$^{1}$ \hspace{0.3cm}
    \bf Ivana Bala{\v z}evi{\' c}\thanks{~~Work done while an intern at Facebook AI Research.}$\:\,^{2}$ \hspace{0.3cm}
    \bf Eric Wallace$^{3}$  \\
    \bf Fabio Petroni$^{4}$ \hspace{0.3cm}
    \bf Sameer Singh$^{1}$ \hspace{0.3cm}
    \bf Sebastian Riedel$^{4,5}$ \hspace{0.3cm}\\
    $^1$UC Irvine \hspace{0.3cm}
    $^2$University of Edinburgh \\
    $^3$UC Berkeley \hspace{0.3cm}
    $^4$Facebook AI Research \hspace{0.3em}
    $^5$University College London \\
    \{\href{mailto:rlogan@uci.edu}{\tt rlogan},\href{mailto:sameer@uci.edu}{\tt sameer}\}\href{mailto:rlogan@uci.edu}{\tt @uci.edu} \hspace{0.3em}
    \href{mailto:ivana.balazevic@ed.ac.uk}{\tt  ivana.balazevic@ed.ac.uk} \\
    \href{mailto:ericwallace@berkeley.edu}{\tt ericwallace@berkeley.edu} \hspace{0.3cm}
    \{\href{mailto:fabiopetroni@fb.com}{\tt fabiopetroni},\href{mailto:sriedel@fb.com}{\tt sriedel}\}\href{mailto:sriedel@fb.com}{@fb.com}
}

\date{}

\begin{document}
\maketitle

\begin{abstract}
Prompting language models (LMs) with training examples and task descriptions has been seen as critical to recent successes in few-shot learning.
In this work, we show that finetuning LMs in the few-shot setting can considerably reduce the need for prompt engineering.
In fact, one can use \emph{null prompts}, prompts that contain neither task-specific templates nor training examples, and achieve competitive accuracy to manually-tuned prompts across a wide range of tasks.
While finetuning LMs does introduce new parameters for each downstream task, we show that this memory overhead can be substantially reduced: finetuning only the bias terms can achieve comparable or better accuracy than standard finetuning while only updating 0.1\% of the parameters.
All in all, we recommend finetuning LMs for few-shot learning as it is more accurate, robust to different prompts, and can be made nearly as efficient as using frozen LMs.
\end{abstract}

\input{sections/10-intro}
\input{sections/20-background}
\input{sections/30-experimental-setup}
\input{sections/40-results}
\input{sections/60-conclusion}
\typeout{}
\bibliography{journal-abbrv,bib}
\bibliographystyle{acl_natbib}

\clearpage
\appendix
\input{sections/99-appendix.tex}

\end{document}

%% file: sections/10-intro.tex
\section{Introduction}\label{sec:intro}

\begin{figure*}[ht]

    \renewcommand\tabularxcolumn[1]{m{#1}}
    \definecolor{t_yellow}{HTML}{92921c}
    \definecolor{t_magenta}{HTML}{ae4388}
    \definecolor{t_green}{HTML}{3c8575}
    \definecolor{t_blue}{HTML}{8692c6}
    \newcommand{\field}[2]{{\{#2\}\textsubscript{#1}}}
    \newcommand{\pattern}[1]{{\color{t_magenta} #1}}
    \newcommand{\pred}[1]{{\color{t_green} #1}}
    
    \begin{minipage}{\textwidth}
        \noindent
        \begin{minipage}{\textwidth}
        \end{minipage}
    
        \centering
        \begin{minipage}{0.35\textwidth}
            \vspace{1.5em} 
            \centering
            \includegraphics{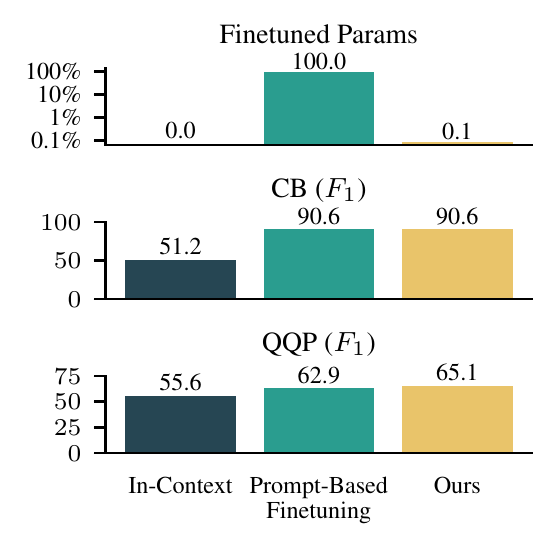}
        \end{minipage}%
        \hfill%
        \begin{minipage}{0.65\textwidth}
            \centering
            \small
            \begin{tabularx}{\textwidth}{ c | X }
                \toprule
                In-Context &
                    \tiny
                    \field{1}{What does it feel like to be on Xanax?} \pattern{and} \field{2}{Do 4mg Xanax bars exist?} \pattern{have} \pred{different} \pattern{meanings.} 
                    \field{1}{How do you know if you're unconditionally in love with someone?} \pattern{and} \field{2}{How do you know if you're in love with someone and might only be denying the fact to yourself?} \pattern{have} \pred{similar} \pattern{meanings.} 
                    \field{1}{Will GST affect the price level in India?} \pattern{and} \field{2}{Will GST effect the price level in India?} \pattern{have} \pred{[MASK]} \pattern{meanings.}
                \\
                \midrule
                \makecell{Prompt-Based \\Finetuning} &
                    \scriptsize 
                    \field{1}{Will GST affect the price level in India?} \pattern{?} \pred{[MASK]} \pattern{, I want to know} \field{2}{Will GST effect the price level in India?}
                \\
                \midrule
                \makecell{Null Prompts \\ (Ours)} & 
                    \footnotesize 
                    \field{1}{Will GST affect the price level in India?} \field{2}{Will GST effect the price level in India?} \pred{[MASK]} \\
                \bottomrule
            \end{tabularx}
        \end{minipage}%
    \end{minipage}
    \vspace{-0.4cm}
    \caption{
        {\bf Different Methods of Few-Shot Learning.} \underline{Right:} We visualize different types of prompts for QQP. We denote the input fields using curly brackets \{\}, the manually-written pattern using \pattern{magenta}, and the verbalizers using \pred{green}.
        We show that \textit{null prompts}, ones that do not contain training examples or task-specific patterns, can achieve competitive accuracy.
        \underline{Left:} We compare different methods for model finetuning. Unlike standard prompt-based finetuning, we propose to update only the masked LM's bias terms (BitFit). This achieves competitive accuracy while only updating 0.1\% of the parameters.
    }
    \label{fig:hook}
\end{figure*}

Few-shot learning---the ability to learn tasks with limited examples---is an important academic and practical challenge~\cite{lake2015human}.
In state-of-the-art NLP, few-shot learning is performed by reformulating tasks as natural language ``prompts’’ and completing those prompts with pre-trained language models~\cite{brown2020language,schick2020exploiting}. 
Prompts that are well-designed can substantially improve accuracy~\cite{zhao2021calibrate,lu2021fantastically}.
However, finding these prompts is difficult: it requires a non-trivial combinatorial search over the prompt's wording (a.k.a. its pattern or template), whether and how to include training examples, and how to convert language model probabilities into class predictions.
Consequently, prompts are often designed using human intuition that is hard to replicate and apply in a principled manner~\cite{perez2021true}.

In this work, we seek to mitigate prompt engineering by identifying a class of simple prompts that are effective across many tasks for masked language models (LMs).
We find that, 
when using prompt-based finetuning~\cite{schick2020exploiting,gao2020making},
the prompt requires less optimization than previously thought; in fact, the pattern and training examples can be completely cut out (e.g., Figure~\ref{fig:hook}, right).
These \emph{null prompts}---simple concatenations of the inputs and the \mask token---achieve comparable accuracy to manually-written patterns while drastically simplifying prompt design: users only need to decide the label names (a.k.a. the verbalizer) and where to place the \mask token.
The effectiveness of null prompts also challenges the common wisdom that the success of few-shot learning is due to inductive biases present in the prompt.

A key drawback of prompt-based finetuning is that it has large memory requirements for each new downstream task (Figure~\ref{fig:hook}, left). 
In contrast, in-context learning~\cite{brown2020language} allows reusing large-scale LMs as is, but it requires significant prompt engineering.
To determine whether memory efficiency and simple prompt selection can be \textit{simultaneously} achieved, we experiment with either: (a) making prompts for in-context learning similarly easy to create, or (b) making prompt-based finetuning more memory efficient.
For (a), we simplify prompt engineering for in-context learning by automatically tuning the prompt's tokens or embeddings, an approach that has been successful in the non-few-shot setting~\cite{shin2020autoprompt,lester2021power}.
For (b), we study lightweight finetuning alternatives that update a smaller set of parameters: BitFit~\cite{bitfit}, Adapters~\cite{houlsby2019parameter}, and calibration layers~\cite{zhao2021calibrate}.

We show that the latter approach---prompt-based finetuning with lightweight updates---is considerably more successful.
In particular, updating only the model's bias terms (BitFit) can achieve competitive or better few-shot accuracy than standard finetuning while only updating 0.1\% of the parameters.
On the other hand, automated prompt tuning for in-context learning generally fails to find prompts that are competitive with manually-engineered ones.
Taken together, our results show that prompt-based finetuning is preferable because it is more accurate, more robust across prompts, and can be made nearly as efficient as using frozen LMs.

%% file: sections/20-background.tex
\section{Prompting Language Models}

\begin{table*}[t]
    \small
    \centering
    \begin{tabular}{lllc}
        \toprule
        \textbf{Method} & \textbf{Finetuned Params} & \textbf{Prompt Design} & \textbf{Few-shot} \\
        \midrule
        \autoprompt~\cite{shin2020autoprompt}   & None & Learned (Discrete) & \xmark \\
        Prompt Tuning~\cite{lester2021power}    & Prompt Token Embeds & Learned (Continuous) & \xmark \\
        \textsc{OptiPrompt}~\cite{zhong2021factual}     & Prompt Token Embeds & Learned (Continuous) & \xmark \\
        Soft Prompts~\cite{qin2021learning}    & All Contextualized Embeds & Learned (Continuous) & \xmark \\
        \midrule
        GPT-3~\cite{brown2020language}          & None & Manual & \cmark\\
        PET~\cite{schick2020exploiting}         & All & Manual & \cmark \\
        LM-BFF~\cite{gao2020making}            & All & Learned (Discrete) & \cmark \\
        P-Tuning~\cite{liu2021gpt}              & All + Prompt Token Embeds & Learned (Continuous) & \cmark \\
        Null Prompts + Bitfit (Ours)            & Bias Terms & None & \cmark \\
        \bottomrule
    \end{tabular}
    \caption{
        {\bf Overview of Existing Work on Prompting}.
        \emph{Finetuned Params} indicates the parameters altered during training. \emph{Prompt Design} indicates how prompts are created; we use \emph{null prompts}. \emph{Few-Shot} indicates using few-shot training and validation sets.
    } 
    \label{tab:background}
\end{table*}

We use masked LMs for few-shot learning. Following \citet{schick2020exploiting}, we have:
\begin{itemize}[nosep,leftmargin=*]
    \item a pre-trained masked LM, with $T$ denoting its token vocabulary and $T^*$ the set of all token sequences.
    \item a small set of training inputs $x_i \in X$ and their corresponding labels $y_i \in Y$.
    \item a \emph{pattern} $P: X \to T^*$ that maps inputs to cloze questions containing a single \texttt{[MASK]} token. Additionally, a \emph{verbalizer} $v: Y \to T$ that maps each label to a single vocabulary token. We call the pattern and verbalizer together the \textit{prompt}.
\end{itemize}

\noindent In our work, we consider different ways of constructing the prompt (Section~\ref{subsec:prompting}) and updating the masked LM's parameters (Section~\ref{subsec:finetuning}). 
Table \ref{tab:background} contains an overview of existing prompting methods and the settings they are evaluated in.

\subsection{Constructing the Prompt}\label{subsec:prompting}

The prompt is important: different prompts can cause accuracy to vary from near chance to near state-of-the-art~\cite{zhao2021calibrate}. This importance, as well as the nontrivial nature of manually tuning the prompt, has led to growing interest in methods for automatic prompt design~\cite{shin2020autoprompt,gao2020making,lu2021fantastically}. 
These methods search for elements such as (1) the text of the pattern, (2) the tokens in the verbalizers, and (3) whether and how training examples are prepended before the test input.
Unfortunately, while automated prompt search can match the accuracy of manual tuning, it introduces its own complexities. For example, the prompts from \citet{gao2020making} achieve comparable results to manually-designed prompts but are found using large generative models and careful validation.

In this paper, we show that prompt-based finetuning (see Section~\ref{subsec:finetuning}) can considerably reduce the importance of the prompt. 
This does not contradict past work---the extreme importance of the prompt is only true when models are \textit{not finetuned}.

\subsection{Finetuning the LM}\label{subsec:finetuning}

\paragraph{In-context Learning} The most well-known strategy for few-shot learning is using a frozen LM~\cite{brown2020language}. This strategy relies solely on \emph{in-context learning} (a.k.a. priming), where the LM learns by conditioning on the prompt rather than updating its parameters. In-context learning is most successful when using very large (e.g., billions of parameters) LMs, as these models better leverage the prompt~\cite{brown2020language}.

\paragraph{Prompt-Based Finetuning} Rather than using frozen LMs, \emph{prompt-based finetuning} methods finetune all of the LM's parameters~\cite{schick2020exploiting,scao2021data,gao2020making}. 
For masked LMs, this is done by constructing training examples that contain a \mask token and finetuning the masked LM to generate the correct verbalizer token in that position.

The main advantage of prompt-based finetuning over in-context learning is that it achieves higher accuracy, especially when the LM is relatively small~\cite{schick2020size}.
The main downside is that the same model can no longer be reused across different tasks, thus reducing memory efficiency. 
In this paper, we will show an additional benefit to prompt-based finetuning---it makes prompt engineering easier.
Moreover, we will show that the memory inefficiency of prompt-based finetuning can be drastically mitigated using lightweight finetuning alternatives.
\citet{scao2021data} concurrently show that different manually-written patterns lead to similar accuracy for prompt-based finetuning.
We take this a step further and show that writing can be avoided entirely; null patterns which merely concatenate the inputs and the [MASK] tokens also have similar accuracy, yet have a substantially simpler design space.

%% file: sections/30-experimental-setup.tex
\section{Experimental Setup}

\subsection{Datasets and Hyperparameter Tuning}\label{subsec:datasets}

We use the following classification datasets from GLUE~\cite{wang2018glue} and SuperGLUE~\cite{wang2019superglue}:
BoolQ, 
CB, 
MNLI, 
MRPC, 
QNLI, 
QQP, 
RTE, 
and SST-2.\footnote{We also evaluated on WiC and WNLI. We omit these results because all models achieved near-random accuracy.}

To build few-shot datasets, past work collects $K$ examples from each label for training and $K$ examples from each label for development~\cite{gao2020making}. Despite this setup often being denoted as $K$-shot learning, it effectively uses $2K$ examples and splits the examples evenly into train and development. We instead propose to use cross validation to perform more principled model selection. Concretely, we sample $2K$ examples from each label and use 4-fold cross validation to determine the best hyperparameters. After finding the best hyperparameters, we train on the first $K$ examples and early stop on the second $K$ examples. We use $K=16$ following past work~\cite{gao2020making}.

We sample our examples from each dataset's original training set. Since few-shot learning can be high variance, we sample the examples with 10 different random seeds and report the mean and variance of the model performance. We use each dataset's original development set for our final evaluation and use the standard evaluation metrics (accuracy or $F_1$) associated with each dataset. We do not check the final evaluation metrics during any tuning of the hyperparameters to ensure that we are doing ``true'' few-shot learning~\cite{perez2021true}.

\begin{figure}[tb]
    \centering
    \includegraphics{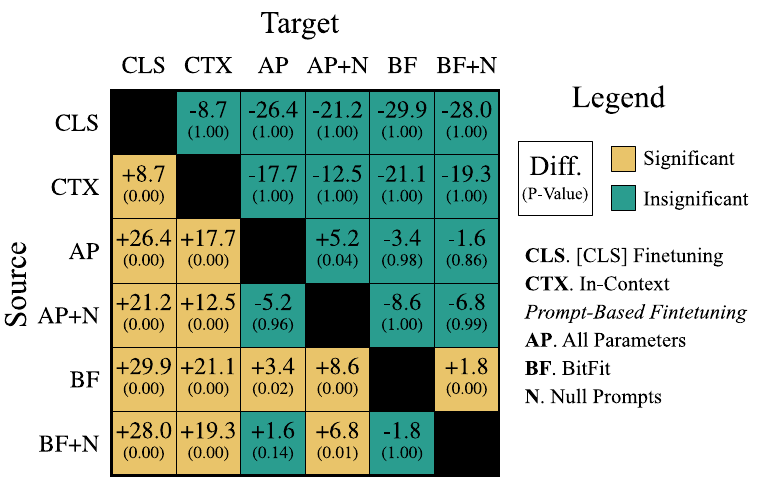}
    \caption{
        {\bf How \wins are Computed}.
        For a given dataset, we perform a Welch's $t$-test to determine if there is a significant difference in accuracy for each pair of methods.
        The method which performs better than most other methods (i.e., the row with the most yellow squares; BitFit in this case) is considered the ``winner'' of the task, and its \emph{\wins} is incremented by 1. In the figure above, we show a subset of methods evaluated on a single dataset. 
    }
    \label{fig:wins}
\end{figure}

\subsection{Masked Language Models}\label{subsec:models}
Following past work~\cite{schick2020size}, we use the RoBERTa~\cite[large, 330M params,][]{liu2019roberta} and ALBERT~\cite[xxl-v2, 223M params,][]{lan2019albert} masked LMs provided by the HuggingFace \texttt{transformers} library~\cite{Wolf2019HuggingFacesTS}.

\subsection{Comparing Few-shot Methods by \wins}
The results for different few-shot learning methods can be quite different across datasets and seeds for the training set~\cite{zhao2021calibrate,schick2020exploiting}. To compare different methods at a high level, we use a metric denoted as \emph{\wins}: the number of datasets that a given method performs significantly better than all other methods on. We compute this metric for a given dataset by first performing a Welch's $t$-test to determine if there is a significant difference in accuracy for each pair of methods. The method which performs better than most other methods is considered the ``winner'' of the task and its \emph{\wins} is incremented by 1. There are multiple winners in the case of a tie. See Figure~\ref{fig:wins} for a demonstration.

%% file: sections/40-results.tex
\begin{figure*}[!htbp]
    \centering
    \includegraphics{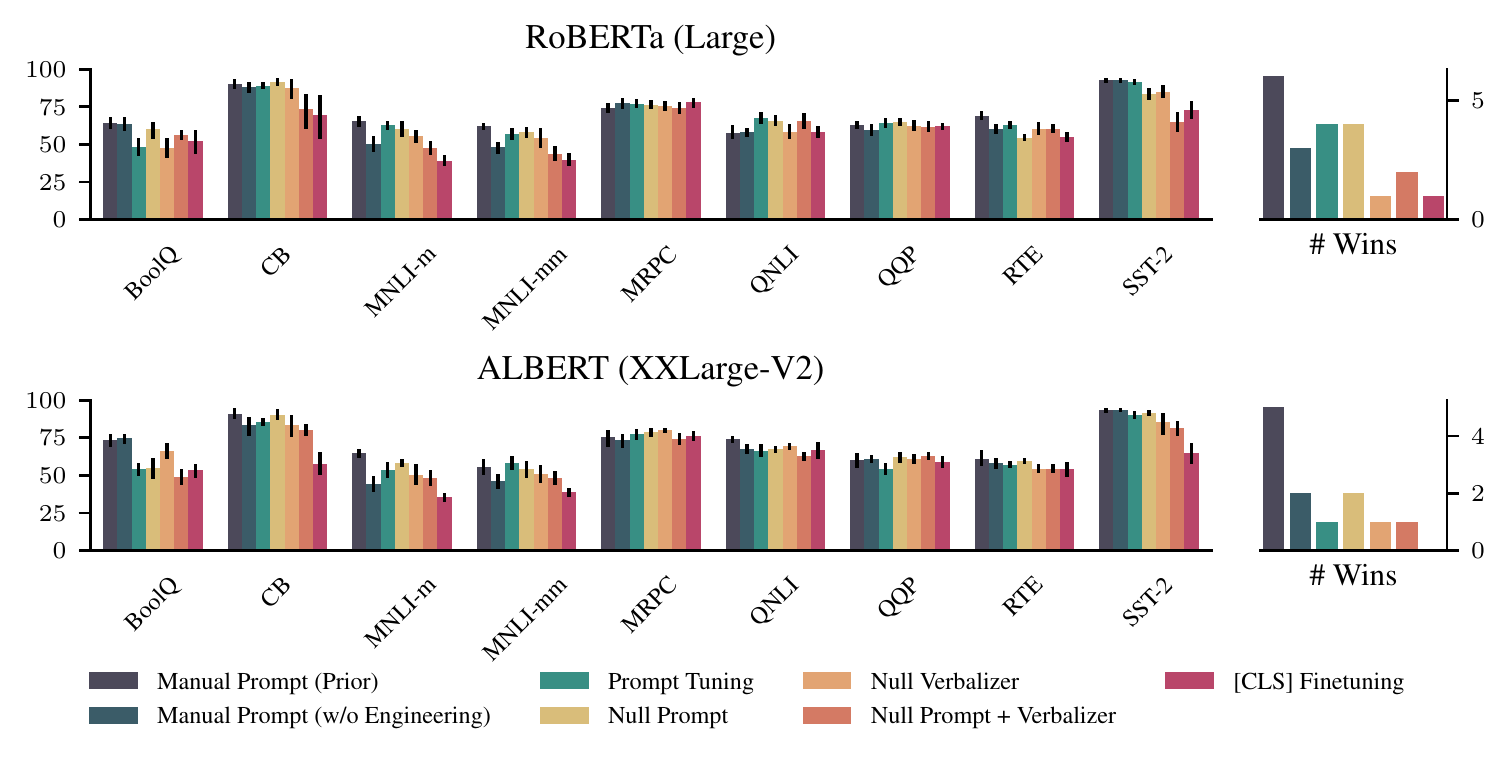}
    \vspace{-0.35cm}
    \caption{
        {\bf Simplifying the Selection of Prompts}.
        We apply prompt-based finetuning in conjunction with six different types of prompts.
        We report accuracy or $F_1$ on each dataset.
        Manually-designed prompts from prior work achieve the best accuracy but require manual tuning on validation sets.
        On the other hand, null prompts and prompt tuning both perform competitively without requiring any tuning of the pattern.
    }
    \label{fig:pvp-ablations}
\end{figure*}

\section{Simplifying Prompt Engineering}\label{sec:pvp}

In this section, we run prompt-based finetuning and ablate different elements of the prompt. We consider the following ablations:
\begin{itemize}[leftmargin=*]
    \item {\bf Manual Prompt (Prior)}: We use manually-written prompts from \citet{schick2020exploiting,schick2020size}. These works do not specify how they obtained these prompts---they may have been tuned using large validation sets. We show the patterns and verbalizers in Appendix~\ref{appendix:prompts}.
    \item {\bf Manual Prompt (w/o Engineering)}: We simulate standard prompt design by manually writing one prompt for each task using our intuition. We show the prompts in Appendix~\ref{appendix:prompts2}.
    \item {\bf Prompt Tuning}: Inspired by \citet{liu2021gpt} and \citet{lester2021power}, we use the pattern from Manual Prompt (Prior) but randomly initialize the embeddings of the pattern tokens and learn them using gradient-based optimization. This ablates the gains from human-designed patterns.
    \item {\bf Null Prompt}: We use the same verbalizer as Manual Prompt (Prior) but use a pattern that consists of only the input fields and a \texttt{[MASK]} token (Appendix~\ref{appendix:prompts3}). This ablates the pattern entirely.
    \item {\bf Null Verbalizer}: We use the same pattern as Manual Prompt (Prior) but select random tokens for the verbalizer. This ablates the gains from a human-designed verbalizer.
    \item {\bf Null Prompt + Verbalizer} We use both null prompts and random tokens for the verbalizer.
\end{itemize}
\vspace{-0.3cm}In all cases, we finetune all of the masked LM parameters. We show the accuracy of the above prompts as well as traditional finetuning (using a \cls token and a classification head) in Figure~\ref{fig:pvp-ablations}.

\paragraph{Manual Prompts Perform Best} The manually-written prompts from prior work perform best on average for both models. However, it is unclear how these prompts were obtained. On the other hand, our manual prompts (w/o Engineering) are noticeably worse than the ones from prior work and are outperformed by many other methods.\smallskip

\noindent \textbf{Null Prompts Are Competitive}
In many cases, prompt tuning and null prompts perform comparably to manually-written prompts, especially for RoBERTa.
For instance, both of these methods outperform our manually-written prompts in terms of \wins.
These results are exciting from a practical perspective as they show that one can achieve competitive few-shot results without resorting to any tuning of the prompt.

From an analysis perspective, these results also show that effective few-shot learning can be accomplished without any inductive bias from a manually-written pattern.
In fact, combining null prompts with null verbalizers, which involves no human design at all, still significantly outperforms standard \cls finetuning for numerous tasks (3 for RoBERTa and 5 for ALBERT at $p=0.05$).
This shows that some of the effectiveness of prompt-based finetuning is due to its basic setup, i.e., predicting on a \mask token with an MLM head.

\paragraph{Null Prompts or Prompt Tuning?}
Both null prompts and prompt tuning achieve competitive results without resorting to manual prompt design.
We advocate for using null prompts over prompt tuning because they are easier to use.
Null prompts only require choosing which order to concatenate the input fields and the \mask token. Prompt tuning requires choosing the number of embeddings, their placement, their initialization, etc. 

\begin{figure}[t]
    \centering
    \includegraphics{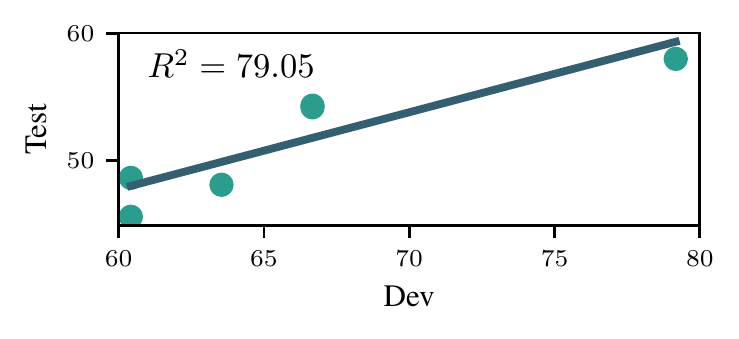}
    \vspace{-0.5cm}
    \caption{
        The only decision to make when using null prompts is which order to concatenate the mask token and the input fields. One can robustly choose the best option using a tiny held-out development set. We show the results for MNLI, with the few-shot development set accuracy on the x-axis.
    }
    \label{fig:mnli-rsquared}
\end{figure}

Moreover, determining the concatenation order for null prompts is trivial by just trying all possible options and choosing which one works best on the validation set.
To see this, in Figure~\ref{fig:mnli-rsquared} we plot the accuracy on the few-shot development set and the full test set for different concatenation orders for RoBERTa on MNLI.\footnote{We use MNLI because the concatenation order has a large impact on performance.}
The development and test accuracy is strongly correlated ($R^2=79.05$), which demonstrates that tuning the concatenation order is easy even when validation data is scarce. In our experiments we use arbitrary concatenation orders; null prompts may more effective with tuning.

\section{Achieving Simplicity \textit{and} Efficiency}\label{sec:efficiency}

{
\begin{figure*}[t]
    \centering
    \includegraphics{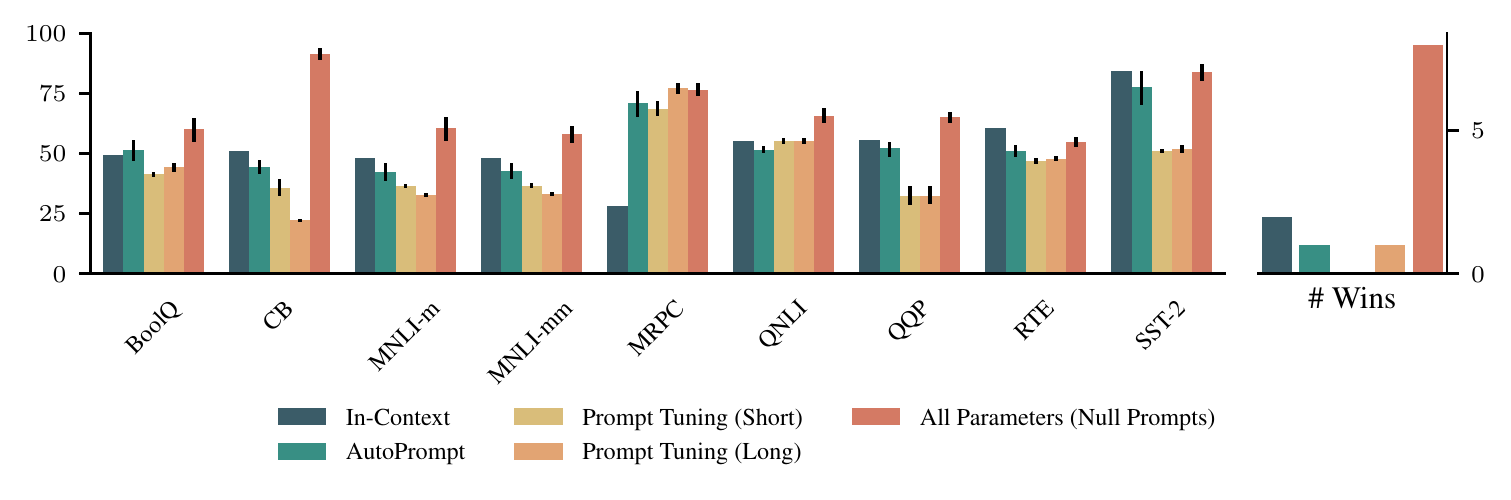}
    \vspace{-0.4cm}
    \caption{
        {\bf Prompt-Only Tuning}. We try to simplify prompt engineering for in-context learning (i.e., using frozen models) by directly learning the prompt. The performance (accuracy/$F_1$) for prompt-only tuning is substantially lower than finetuning the LM parameters for RoBERTa-large. Thus, we recommend finetuning over in-context learning in the few-shot setting.
    }
    \label{fig:prompt-tuning}
\end{figure*}

\begin{figure*}[t]
    \centering
    \includegraphics{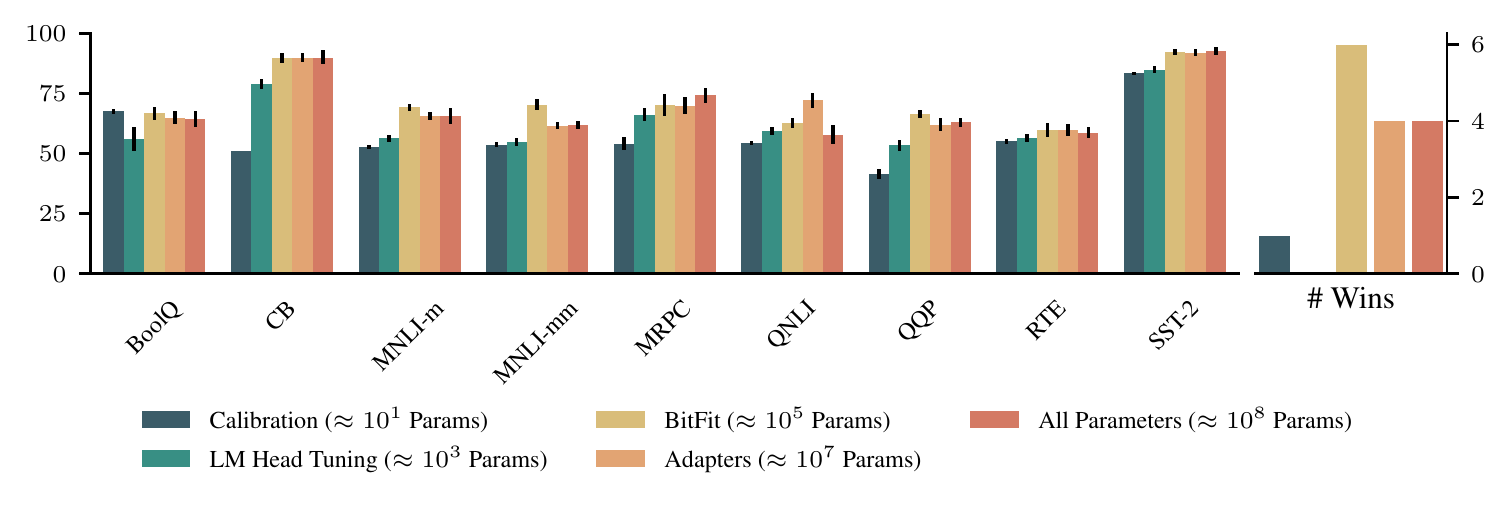}
    \vspace{-0.4cm}
    \caption{
        {\bf Parameter-Efficient Prompt-based Finetuning}. We perform prompt-based finetuning using different lightweight finetuning schemes. We show the accuracy or $F_1$ on each dataset for RoBERTa-large. BitFit achieves the highest accuracy on average and only modifies 0.1\% of the parameters.
    }
    \label{fig:parameter-tuning}
\end{figure*}
}

Thus far, we have shown that prompt-based finetuning can simplify prompt engineering at the cost of memory inefficiency---a new set of parameters must be learned for each task. This is in contrast to in-context learning, which holds all model weights fixed but is heavily influenced by small prompt modifications~\cite{zhao2021calibrate,lu2021fantastically}.
In this section, we investigate how to achieve \textit{both} memory efficiency and simple prompts.
Concretely, in Section~\ref{subsec:prompt-efficient} we try to simplify prompt engineering for in-context learning by tuning the prompt, and in Section~\ref{subsec:parameter-efficiecnt}, we reduce the number of learned parameters for prompt-based finetuning.

\subsection{Simplifying In-Context Learning With Prompt-Only Tuning}\label{subsec:prompt-efficient}

Here, we try to make prompt engineering for in-context learning as simple as prompt-based finetuning by automatically finding the prompt. Concretely, we focus on the emerging class of methods that do \textit{prompt-only tuning}: learning the prompt while keeping the rest of the model fixed~\cite{shin2020autoprompt,lester2021power}. We consider:
\begin{itemize}[leftmargin=*,itemsep=0mm]
    \item {\bf \autoprompt}: Following~\cite{shin2020autoprompt}, we search for discrete tokens to use in the input instead of manually-designed patterns. We use the hyperparameters from \citet{shin2020autoprompt}.
    \item {\bf Prompt Tuning (Short)}: We use the same prompt tuning approach described in the previous section but we keep the masked LM fixed.
    \item {\bf Prompt Tuning (Long)}: We increase the number of learned prompt embeddings to 20 in order to expand the learning capacity.
\end{itemize}
For reference, we also report the results from prompt-based finetuning with null prompts.
We show the results for RoBERTa in Figure~\ref{fig:prompt-tuning}.
We find that only tuning the prompt is relatively unsuccessful. First, on average it fails to match the performance of manually-designed prompts. Second, all methods struggle to match the accuracy of prompt-based finetuning. In fact, for many of the datasets, prompt-only methods perform worse by a wide margin (e.g., ~40\% absolute difference in $F_1$ score on CB). This shows that finetuning masked LMs in the few-shot setting leads to substantially higher accuracy than prompt-only tuning.\smallskip

\input{sections/41-table}

\noindent \textbf{Our Results versus Recent Prompt Tuning Work} We find that only tuning the prompt performs substantially worse than finetuning the entire LM. This is in contrast to recent work, which argues that prompt-only tuning is competitive with finetuning~\cite{lester2021power,li2021prefix}. We believe these are not contradictions but rather differences in the models and settings. \citet{li2021prefix} focus on \textit{left-to-right} LMs for \textit{generation} tasks, whereas we focus on masked LMs for classification tasks. This may explain the difference in the results. Moreover, \citet{lester2021power} show that prompt-only tuning becomes less competitive as models get smaller; we use even smaller models than evaluated in their work. Consequently, although we find that finetuning a masked LM is superior to prompt-only tuning, there may be other settings in which they fair similarly.

\subsection{Memory-Efficient Finetuning}\label{subsec:parameter-efficiecnt}

Given the inadequacies of prompt-only tuning, we next study if prompt-based finetuning can be made memory-efficient. To do so, we focus on reducing the number of trainable parameters, taking inspiration from recent work in the non-few-shot setting. We consider four methods:
\begin{itemize}[leftmargin=*,itemsep=0mm]
    \item {\bf Adapters}: We use Adapters~\cite{houlsby2019parameter}, neural networks layers that are inserted between the feedforward portion of the Transformer architecture. We use the default Adapters hyperparameters from \citet{houlsby2019parameter} ($\approx10^7$ parameters per task).
    \item {\bf BitFit}: Following \citet{bitfit}, we only update the bias terms inside the Transformer ($\approx10^5$ parameters per task).
    \item {\bf LM Head Tuning}: We update the embeddings in the MLM output layer that are associated with the verbalizer tokens ($\approx10^3$ parameters per task).
    \item {\bf Calibration}: Following \citet{zhao2021calibrate}, we learn an affine transformation on top of the logits associated with the verbalizer tokens ($\approx10^1$ parameters per task).
\end{itemize}
We run prompt-based finetuning for each method with the prompts from Manual Prompts (Prior). We also report the accuracy of finetuning all of the parameters for reference.

\paragraph{Results} We show the results in Figure~\ref{fig:parameter-tuning}. There are diminishing returns as the parameter count is increased. In particular, substantial gains are made when going from calibration to LM head tuning to BitFit, however, there is either a marginal improvement or even a decrease in performance when going to Adapters or All Parameters.
The BitFit method provides the best accuracy-efficiency trade-off, and it even outperforms finetuning all of the parameters in terms of \wins. This suggests that updating all of the LM's hundreds of millions of parameters on only 16 data points is suboptimal.

\subsection{Putting Everything Together} 
We finally combine null prompts and memory-efficient finetuning. We show the results from this method, as well as the other best few-shot methods, in Table~\ref{table:resultsrob}. Overall, we recommend finetuning with null prompts and BitFit: it achieves competitive accuracy, is simple to set up, and introduces small memory costs for each new task.

%% file: sections/41-table.tex
\begin{table*}[!t]
	\centering
	\footnotesize
    \begin{tabular}{cl|cccccccc|c}
    \toprule 
        & & BoolQ & CB & MNLI & MRPC & QNLI & QQP & RTE & SST-2 & Wins \\
        & & (acc) & ($F_1$) & (acc) & ($F_1$) & (acc) & ($F_1$) & (acc) & (acc) & (\#)\\
    \midrule 
    \multirow{7}{*}{\STAB{\rotatebox[origin=c]{90}{RoBERTa}}}
    & In-context          &     49.2 &     51.2 &     48.0 / 48.1 &     28.0 &     55.2 &     55.6 &     60.7 &     84.1 & 0 \\
    & \cls finetuning             &     51.0 &     74.3 &     39.4 / 38.6 & \bf 77.8 &     58.2 &     61.9 &     54.5 &     72.9 & 1 \\
    & \textit{Prompt-based Finetuning} & & & & & & & & &\\

    & \hspace{0.4em} All Parameters &     63.9 & \bf 90.6 &     66.5 / 61.6 &     74.1 &     57.4 &     62.9 & \bf 68.8 & \bf 92.6 & 3 \\
    & \hspace{1.2em} + Null Prompt   &     59.9 & \bf 91.2 &     61.6 / 57.8 & \bf 76.1 &     65.8 & \bf 65.9 &     54.6 &     83.8 & 3 \\
    & \hspace{0.4em} BitFit & \bf 66.7 & \bf 89.8 & \bf 69.3 / \bf 70.0 &     69.7 &     62.3 & \bf 66.3 &     64.9 & \bf 92.1 & 6 \\
    & \hspace{1.2em} + Null Prompt    & \bf 67.2 & \bf 90.6 &     67.5 / 62.9 &     68.2 & \bf 66.4 &     65.1 &     65.4 &     89.6 & 3 \\
    \midrule
    \multirow{7}{*}{\STAB{\rotatebox[origin=c]{90}{ALBERT}}}

    & In-context         &     68.0 &     19.9 &     35.4 / 35.2 &     20.7 &     50.1 &     \phantom{0}0.3 &     53.1 &     49.1 & 0 \\
    & \cls finetuning             &     53.3 &     56.5 &     36.0 / 38.6 & \bf 76.9 &     66.6 & \bf 58.5 &     54.1 &     62.9 & 2 \\
    & \textit{Prompt-based Finetuning} & & & & & & & & &\\
    & \hspace{0.4em} All Parameters & \bf 73.5 & \bf 91.1 & \bf 65.0 / 56.0 & \bf 75.2 & \bf 73.9 & \bf 59.9 & \bf 61.4 & \bf 93.2 & 8 \\
    & \hspace{1.2em} + Null Prompt  &     53.7 & \bf 89.4 &     58.2 / 53.7 & \bf 78.5 &     67.3 & \bf 62.0 &     59.2 &     91.5 & 3 \\
    & \hspace{0.4em} BitFit & \bf 77.2 & \bf 86.7 & \bf 64.6 / \bf 61.6 & \bf 79.7 & \bf 73.1 & \bf 61.4 &     58.6 & \bf 92.0 & 8 \\
    & \hspace{1.2em} + Null Prompt  &     52.8 & \bf 86.3 &     55.3 / 58.0 &     65.5 &     63.8 &     52.7 &     57.2 &     89.7 & 1 \\
    \bottomrule
    \end{tabular}
    \caption{
        {\bf Final Few-shot Results} from representative methods. Wins are computed on a per-datasets basis and the ``winners'' of the different approaches are highlighted in bold. Prompt-based finetuning significantly outperforms in-context learning and traditional \cls finetuning, even without any tuning of the prompt (\textit{null prompt}). Moreover, prompt-based finetuning can be highly memory efficient using bias-only finetuning (\textit{BitFit}). We show matched and mismatched results for MNLI.
    }
    \label{table:resultsrob}
 \end{table*}

%% file: sections/60-conclusion.tex
\section{Conclusion and Future Work}

Two high-level methods exist in few-shot prompting: using a frozen LM (in-context learning) and finetuning the LM on the few training examples (prompt-based finetuning). In this work, we demonstrate two new advantages of prompt-based finetuning. First, we show that it is robust to different choices of the prompt. In fact, there is a simple class of prompts---null prompts---that can be flexibly applied to different tasks without degrading performance relative to manually-written and learned prompts. 
Second, we demonstrate that prompt-based finetuning can be made memory efficient: finetuning only the bias terms (BitFit) achieves comparable or better accuracy than finetuning all the parameters while being 1000x more memory efficient. Taken together, using null patterns with BitFit is an approach that is efficient, simple-to-tune, and competitive in accuracy.

Our results motivate future \textit{analysis} of few-shot learning methods. Concretely, we show that the success of prompt-based finetuning is not solely explained by carefully-chosen patterns or verbalizers. This suggests that the gains from prompt-based finetuning are partially due to its low-level setup, i.e., predicting on a \mask token with a pre-trained MLM head. More generally, we hope to further analyze why and how small changes to different few-shot learning methods can lead to wildly different accuracies. 
We also hope to extend our findings to both \textit{very large} and \textit{left-to-right} LMs, as our current results are for masked LMs that are relatively small by modern standards.

%% file: sections/99-appendix.tex
\appendix
\setcounter{table}{0}
\renewcommand{\thetable}{A\arabic{table}}

\newcolumntype{b}{X}
\newcolumntype{s}{>{\hsize=.33\hsize}X}
\begin{table*}[t!]
    \small
    \centering
    \begin{tabularx}{\linewidth}{rbs}
        \toprule
        \textbf{Dataset} & \textbf{Pattern} & \textbf{Verbalizer} \\ 
        \midrule
        \multirow{2}{*}{BoolQ}   & \multirow{2}{*}{\{passage\}. Question: \{question\}? Answer: [MASK].}
                & True: "Yes" \\
                & & False: "No" \\
        \midrule
        \multirow{3}{*}{CB}      & \multirow{3}{*}{\{premise\}? [SEP] [MASK], \{hypothesis\}}
                & entailment: "Yes"\\
                & & contradiction: "No" \\
                & & neutral: "Maybe" \\
        \midrule
        \multirow{3}{*}{MNLI}    & \multirow{3}{*}{\{sentence1\}? [SEP] [MASK], \{sentence2\}}
                & entailment: "Yes" \\
                & &  contradiction: "No" \\
                & &  neutral: "Maybe" \\
        \midrule
        \multirow{3}{*}{MNLI-mm} & \multirow{3}{*}{\{sentence1\}? [SEP] [MASK], \{sentence2\}}
                & entailment: "Yes" \\
                & &  contradiction: "No" \\
                & &  neutral: "Maybe" \\
        \midrule
        \multirow{2}{*}{MRPC}    & \multirow{2}{*}{\{sentence1\} and \{sentence2\} have [MASK] meanings.}
                & 0: "different" \\
                & &  1: "similar" \\
        \midrule
        \multirow{2}{*}{QNLI}    & \multirow{2}{*}{\{question\}? [SEP] [MASK], \{sentence\}}
                & entailment: "Yes" \\
                & &  not\_entailment: "No" \\
        \midrule
        \multirow{2}{*}{QQP}    & \multirow{2}{*}{\{question1\} and \{question2\} have [MASK] meanings.}
                & 0: "different" \\
                & &  1: "similar" \\
        \midrule
        \multirow{2}{*}{RTE}     & \multirow{2}{*}{\{sentence1\}? [SEP] [MASK], \{sentence2\}}
                & entailment: "Yes" \\
                & &  not\_entailment: "No" \\
        \midrule
        \multirow{2}{*}{SST-2}   & \multirow{2}{*}{\{sentence\} It was [MASK] .}
                & 0: "terrible" \\
                & &  1: "great" \\
        \midrule
    \end{tabularx}
    \vspace{-0.2cm}
    \caption{
        \textbf{Prompts denoted as ``Manual Prompts (Prior)''}. We use prompts inspired from past work~\cite{schick2020exploiting,gao2020making}. The fields between curly brackets indicate dataset-specific inputs. Predictions are made on the \mask token in each prompt.
        For prompt tuning, we tune the tokens in the pattern.
    }
    \label{appendix:prompts}
\end{table*}

\newcolumntype{b}{X}
\newcolumntype{s}{>{\hsize=.33\hsize}X}
\begin{table*}[t!]
    \small
    \centering
    \begin{tabularx}{\linewidth}{rbs}
        \toprule
        \textbf{Dataset} & \textbf{Pattern} & \textbf{Verbalizer} \\ 
        \midrule
        \multirow{2}{*}{BoolQ}   & \multirow{2}{*}{Passage: \{passage\} Question: \{question\} Answer: [MASK].}
                & True: "true" \\
                & & False: "false" \\
        \midrule
        \multirow{3}{*}{CB}      & \multirow{3}{*}{Premise: \{premise\} Hypothesis: \{hypothesis\} Label: [MASK]}
                & entailment: "yes"\\
                & & contradiction: "no" \\
                & & neutral: "maybe" \\
        \midrule
        \multirow{3}{*}{MNLI}    & \multirow{3}{*}{Premise: \{sentence1\} Hypothesis: \{sentence2\} Label: [MASK]}
                & entailment: "yes" \\
                & &  contradiction: "no" \\
                & &  neutral: "maybe" \\
        \midrule
        \multirow{3}{*}{MNLI-mm}    & \multirow{3}{*}{Premise: \{sentence1\} Hypothesis: \{sentence2\} Label: [MASK]}
                & entailment: "yes" \\
                & &  contradiction: "no" \\
                & &  neutral: "maybe" \\
        \midrule
        \multirow{2}{*}{MRPC}    & \multirow{2}{*}{\{sentence1\} and \{sentence2\} are the [MASK].}
                & 0: "different" \\
                & &  1: "same" \\
        \midrule
        \multirow{2}{*}{QNLI}    & \multirow{2}{*}{Question: \{question\} Sentence: \{sentence\} Label: [MASK]}
                & entailment: "yes" \\
                & &  not\_entailment: "no" \\
        \midrule
        \multirow{2}{*}{QQP}    & \multirow{2}{*}{\{question1\} and \{question2\} are the [MASK].}
                & 0: "different" \\
                & &  1: "same" \\
        \midrule
        \multirow{2}{*}{RTE}    & \multirow{2}{*}{Premise: \{sentence1\} Hypothesis: \{sentence2\} Label: [MASK]}
                & entailment: "yes" \\
                & &  not\_entailment: "no" \\
        \midrule
        \multirow{2}{*}{SST-2}   & \multirow{2}{*}{\{sentence\} Overall my impression is [MASK] .}
                & 0: "bad" \\
                & &  1: "good" \\
        \midrule
    \end{tabularx}
    \vspace{-0.2cm}
    \caption{
        \textbf{Prompts denoted as ``Manual Prompts (w/o Engineering)''}. We manually write one prompt for each task, using only our intuition, and do not tune or edit them in any way after evaluating them. Fields between curly brackets indicate dataset-specific inputs.
        Predictions are made on the \mask token in each prompt.
        For prompt tuning, we tune the tokens in the pattern.
    }
    \label{appendix:prompts2}
\end{table*}

\newcolumntype{b}{X}
\newcolumntype{s}{>{\hsize=.33\hsize}X}
\begin{table*}[t!]
    \small
    \centering
    \begin{tabularx}{\linewidth}{rbs}
        \toprule
        \textbf{Dataset} & \textbf{Pattern} & \textbf{Verbalizer} \\ 
        \midrule
        \multirow{2}{*}{BoolQ}   & \multirow{2}{*}{\{passage\} \{question\} [MASK]}
                & True: "Yes" \\
                & & False: "No" \\
        \midrule
        \multirow{3}{*}{CB}      & \multirow{3}{*}{\{premise\} [MASK] \{hypothesis\}}
                & entailment: "Yes"\\
                & & contradiction: "No" \\
                & & neutral: "Maybe" \\
        \midrule
        \multirow{3}{*}{MNLI}    & \multirow{3}{*}{\{sentence1\} [MASK] \{sentence2\}}
                & entailment: "Yes" \\
                & &  contradiction: "No" \\
                & &  neutral: "Maybe" \\
        \midrule
        \multirow{3}{*}{MNLI-mm} & \multirow{3}{*}{\{sentence1\} [MASK] \{sentence2\}}
                & entailment: "Yes" \\
                & &  contradiction: "No" \\
                & &  neutral: "Maybe" \\
        \midrule
        \multirow{2}{*}{MRPC}    & \multirow{2}{*}{\{sentence1\} \{sentence2\} [MASK]}
                & 0: "different" \\
                & &  1: "similar" \\
        \midrule
        \multirow{2}{*}{QNLI}    & \multirow{2}{*}{\{question\} [MASK] \{sentence\}}
                & entailment: "Yes" \\
                & &  not\_entailment: "No" \\
        \midrule
        \multirow{2}{*}{QQP}    & \multirow{2}{*}{\{question1\} \{question2\} [MASK]}
                & 0: "different" \\
                & &  1: "similar" \\
        \midrule
        \multirow{2}{*}{RTE}     & \multirow{2}{*}{\{sentence1\} [MASK] \{sentence2\}}
                & entailment: "Yes" \\
                & &  not\_entailment: "No" \\
        \midrule
        \multirow{2}{*}{SST-2}   & \multirow{2}{*}{\{sentence\} [MASK]}
                & 0: "terrible" \\
                & &  1: "great" \\
        \midrule
    \end{tabularx}
    \vspace{-0.2cm}
    \caption{
        \textbf{Null Prompts} used for results in Sections~\ref{sec:pvp} and~\ref{sec:efficiency}.
    }
    \label{appendix:prompts3}
\end{table*}